\documentclass[letterpaper, 10 pt, conference]{ieeeconf}
\IEEEoverridecommandlockouts
\IEEEoverridecommandlockouts
\usepackage{dblfloatfix}
\usepackage{amsmath,amssymb,amsfonts}
\usepackage{algorithmic}
\usepackage{graphicx}
\usepackage{textcomp}
\usepackage{xcolor}
\usepackage{array}
\usepackage{multirow}
\usepackage{subcaption}
\usepackage{booktabs}
\usepackage{siunitx}
\usepackage[font=small]{caption}

\usepackage{hyperref}

\title{\LARGE \bf{Pose-Anchored Optical Flow for Low-Latency Human Action Anticipation in Human-Robot Teaming
}}

\author{Lewis de Zoete Grundy$^{1}$, Chris McCarthy$^{1}$, and Christopher Fluke$^{1}$
\thanks{$^{1}$School of Science, Computing and Emerging Technologies,
        Swinburne University of Technology, John St, Hawthorn VIC 3122, Australia
        {\tt\small\{ldezoetegrundy,cdmccarthy,cfluke\}@swin.edu.au}}%
}

\usepackage{tikz}
\usetikzlibrary{calc}

\newcommand{\preprintnotice}{%
  \begin{tikzpicture}[remember picture, overlay]
    \node[anchor=south west, align=left, font=\footnotesize,
          text width=0.9\textwidth] at ($(current page.south west)+(0.8in,0.15in)$) {
      This is the accepted version of a paper accepted for publication at
      IEEE RO-MAN 2026, Kitakyushu, Japan. \copyright~2026 IEEE. Personal use of
      this material is permitted. Permission from IEEE must be obtained for
      all other uses, in any current or future media, including
      reprinting/republishing this material for advertising or promotional
      purposes, creating new collective works, for resale or redistribution
      to servers or lists, or reuse of any copyrighted component of this
      work in other works.
    };
  \end{tikzpicture}
}

\begin{document}

\maketitle
\preprintnotice
\thispagestyle{empty}
\pagestyle{empty}

\begin{abstract}
 Human-robot interaction (HRI) requires robots to interpret human actions early in their execution in order to respond safely, efficiently, and naturally. However, many existing approaches to human action recognition rely either on sparse skeletal representations, which lack fine-grained motion cues, or dense optical flow, which can be computationally expensive for low-latency perception pipelines. In this paper, we propose PoseOFF, a pose-anchored optical flow representation that captures local motion information around human joints to support earlier human intent understanding. By conditioning motion feature extraction on human pose, PoseOFF encodes localised motion dynamics at semantically meaningful body locations, forming a structured motion representation that is explicitly aligned with human kinematics. We evaluate PoseOFF across multiple benchmark datasets and backbone architectures for action anticipation, demonstrating consistent improvements in recognition accuracy, particularly at early observation ratios. Our results show that PoseOFF enables models to achieve comparable or improved performance while observing less of the action sequence, highlighting its effectiveness for early prediction. Importantly, these gains are achieved without requiring full-frame motion processing, making the approach practical for real-time and resource-constrained settings. These findings suggest that pose-centred motion representations such as PoseOFF can enhance the ability of interactive robot systems to infer human actions earlier, supporting more responsive and anticipatory behaviour in human-robot interaction scenarios.
\end{abstract}

\section{Introduction}
Effective human-robot interaction (HRI) relies on a robot’s ability to interpret human behaviour as it unfolds, rather than after it has completed. In collaborative environments, robots must anticipate human actions to coordinate movements, avoid interference, and ensure safety in shared workspaces \cite{lasota2017SurveyMethodsSafeHumanRobotInteraction,villani2018SurveyhumanrobotcollaborationindustrialsettingsSafetyintuitiveinterfacesapplications}. In assistive and social contexts, early recognition of user intent enables timely support and more natural interaction, supporting proactive assistance and fluid turn-taking. Across these domains, the ability to infer human intent from partial observations is central to responsive and human-aware robotic systems.

Despite significant progress in human action recognition, early action anticipation remains a challenging problem. Many approaches rely on skeletal representations derived from pose estimation, which provide a compact and semantically meaningful description of human body configuration \cite{yan2018SpatialTemporalGraphConvolutionalNetworksSkeletonBasedActionRecognition,shi2019SkeletonBasedActionRecognitionDirectedGraphNeuralNetworks}. These representations are well suited to real-time systems due to their efficiency and robustness to background variation. However, skeleton-based methods inherently discard fine-grained motion information, limiting their ability to distinguish between actions with similar pose configurations but different movement dynamics, particularly in the early stages of an action.

Conversely, dense motion representations such as optical flow capture rich spatiotemporal information and have been widely used to improve action recognition performance \cite{simonyan2014TwoStreamConvolutionalNetworksActionRecognitionVideos}. More recent work has explored transformer-based architectures for action anticipation, achieving strong performance through large-scale spatiotemporal modelling \cite{girdhar2021AnticipativeVideoTransformer}. However, these approaches often come with increased computational cost, making them less suitable for low-latency perception pipelines required in interactive robotic systems.

\begin{figure}[t]
    \centering
    \includegraphics[width=0.8\linewidth]{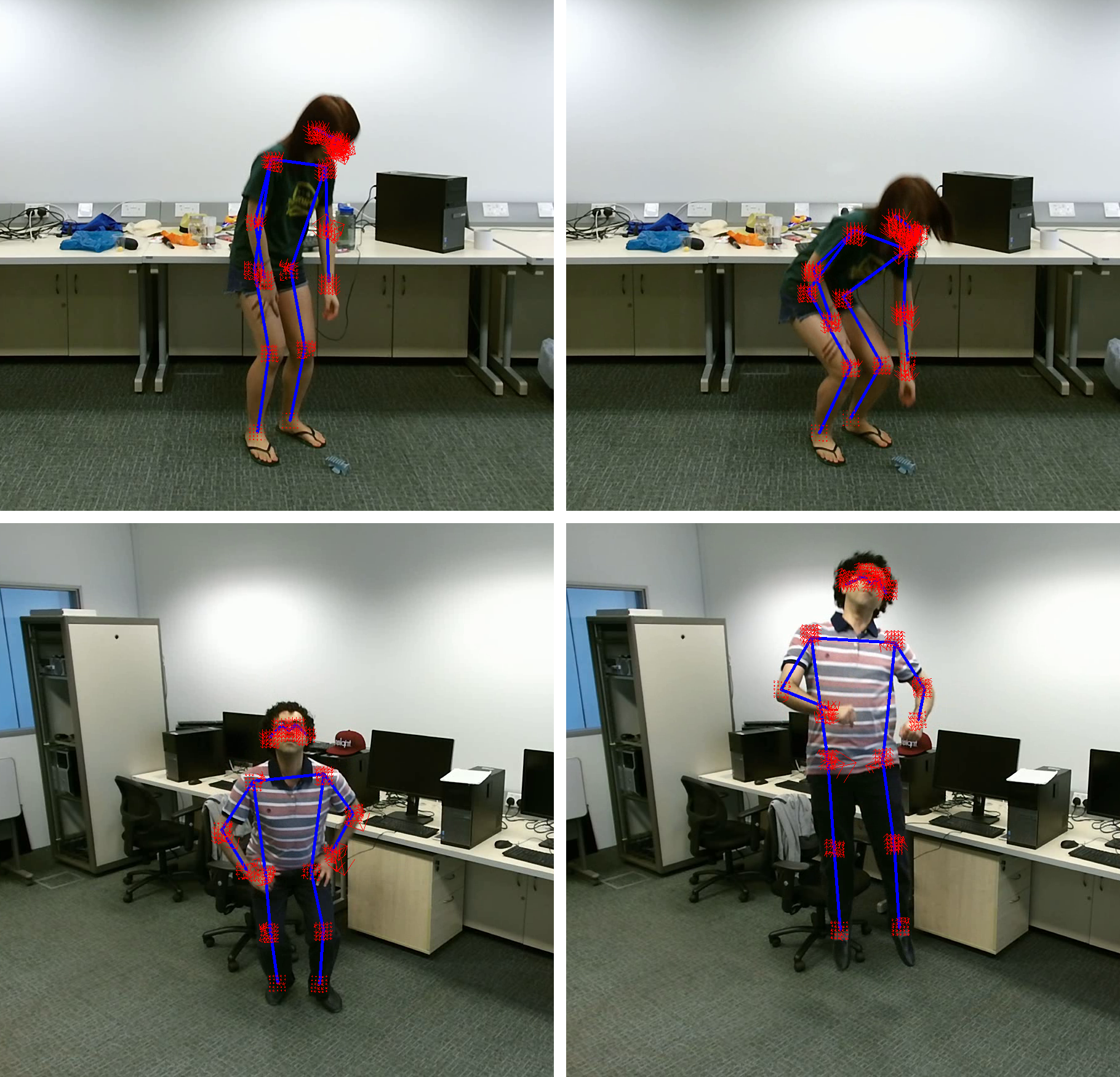}
    \caption{PoseOFF captures localised motion patterns (e.g., rotation, divergence) around key joints that are not represented in skeletal pose alone.}
    \label{fig:classexamples_small}
\end{figure}

This trade-off between efficiency and motion fidelity presents a key challenge for deploying action anticipation methods in HRI settings. A promising direction is to focus on local, body-centred motion cues that retain informative dynamics while avoiding the cost of full-frame processing. By anchoring motion representations to human pose keypoints, it is possible to capture the most relevant movement patterns associated with human actions, while maintaining a compact and interpretable representation. Such approaches align well with the requirements of HRI systems, where perception modules must operate efficiently and provide timely insights into human behaviour.

In HRI settings, earlier recognition is often more valuable than marginal gains in full-sequence classification accuracy because it directly affects response timing, coordination quality, and safety margins. Prior work in HRI has shown that the timing and predictability of robot responses play a critical role in interaction fluency and human trust \cite{dragan2013Legibilitypredictabilityrobotmotion}. A method that can reliably infer human actions from partial observations—even with slightly lower peak accuracy—may therefore be more useful in practice than one that performs best only after the action has fully unfolded.

While prior work has explored combining pose and motion information, these approaches typically rely on either parallel processing streams or global motion representations, without explicitly structuring motion features around human kinematics. As a result, motion information is often treated as a dense signal rather than a behaviourally grounded representation aligned with the human body. This limits both interpretability and efficiency, particularly in early action anticipation, where informative motion cues are localised and temporally sparse.

Motivated by this, we propose PoseOFF, a pose-conditioned motion representation that encodes localised optical flow features at human joint locations that captures local motion information around human joints to support early action anticipation. PoseOFF samples optical flow in the vicinity of estimated keypoints, producing a compact description of motion that complements skeleton-based representations. Importantly, this representation can be integrated into existing architectures with minimal modification, providing a lightweight mechanism for enhancing early human intent understanding in interactive robot systems.

We evaluate PoseOFF across multiple benchmark datasets and backbone models for action anticipation, demonstrating consistent improvements in performance at early observation ratios. Our results show that incorporating local motion cues enables models to achieve comparable or improved recognition accuracy while observing less of the action sequence. These findings highlight the potential of pose-centred motion representations as a practical enhancement for low-latency perception pipelines in HRI applications.

In summary, this paper makes the following contributions:
\begin{itemize}
    \item We propose PoseOFF, a novel pose-conditioned motion representation that anchors optical flow to human joints, introducing a structured, body-aligned alternative to dense or parallel motion representations.

    \item We show that this representation enables earlier and more discriminative action anticipation, achieving comparable accuracy with reduced observation.

    \item We demonstrate the value of targeted, low-latency motion representations for HRI, supporting earlier and more responsive human–robot interaction.
\end{itemize}
This paper is structured as follows: Section~\ref{sec:related_work} reviews related work. Section~\ref{sec:method} presents the PoseOFF representation. Section~\ref{sec:experiments} reports experimental results, including action anticipation, per-class analysis, and performance considerations. Section~\ref{sec:discussion_conclusion} concludes with discussion and future directions.

\begin{figure*}[ht]
    \centering
    \includegraphics[width=0.9\linewidth]{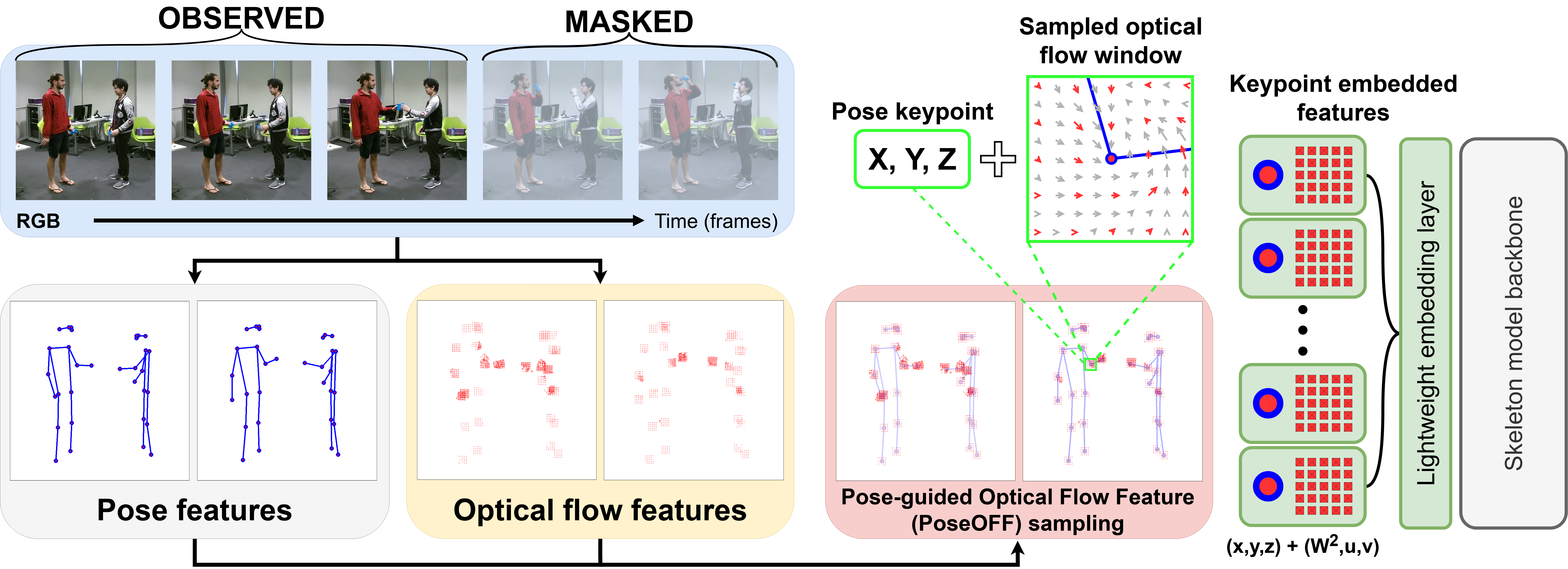}
    \caption{\label{fig:poseoff_extraction}
    PoseOFF representation construction. From an input RGB video, human pose keypoints and optical flow fields are extracted. Motion features are computed at each joint location by encoding local optical flow dynamics in the vicinity of the keypoint. These pose-conditioned motion features are combined with joint coordinates to form a structured representation aligned with human kinematics, capturing localised motion information for each body part. Motion feature windows are concatenated with keypoint information without alteration to skeleton joint relations. A lightweight embedding layer allows the PoseOFF data to be fed into an existing skeleton-based action recognition models without modification to model backbone.
    }
\end{figure*}

\section{Related work}
\label{sec:related_work}

\subsection{Human Action Anticipation for Interactive Systems}

Human action anticipation aims to recognise actions from partial observations, enabling early inference of human intent. This capability is particularly important in interactive robot systems, where timely interpretation of human behaviour supports safe collaboration and responsive interaction \cite{dragan2013Legibilitypredictabilityrobotmotion,lasota2017SurveyMethodsSafeHumanRobotInteraction}.

Early approaches relied on probabilistic models and handcrafted features \cite{ryoo2011HumanactivitypredictionEarlyrecognitionongoingactivitiesstreamingvideos}, while later work introduced recurrent architectures to encourage prediction from partial sequences \cite{aliakbarian2017EncouragingLSTMsAnticipateActionsVeryEarly}. More recently, transformer-based models such as AVT \cite{girdhar2021AnticipativeVideoTransformer} have demonstrated strong performance by modelling long-range temporal dependencies, with multimodal approaches further incorporating complementary cues \cite{kong2022HumanActionRecognitionPredictionSurvey}. However, these methods typically rely on dense spatiotemporal representations and large models, limiting their suitability for real-time deployment in interactive robotic systems.

\subsection{Skeleton-based Human Action Recognition}

Skeleton-based representations provide a compact and semantically meaningful description of human motion through joint trajectories, and are widely used due to their efficiency and robustness to background variation \cite{yan2018SpatialTemporalGraphConvolutionalNetworksSkeletonBasedActionRecognition,shi2019SkeletonBasedActionRecognitionDirectedGraphNeuralNetworks}. Graph-based approaches such as ST-GCN \cite{yan2018SpatialTemporalGraphConvolutionalNetworksSkeletonBasedActionRecognition} model spatial and temporal relationships between joints, while extensions such as 2s-AGCN \cite{shi2019SkeletonBasedActionRecognitionDirectedGraphNeuralNetworks} introduce adaptive graph structures and motion streams to improve representation capacity. More recent methods, including PoseC3D \cite{duan2022RevisitingSkeletonbasedActionRecognition}, explore alternative formulations using spatiotemporal convolutions over pose sequences.

These approaches achieve strong performance while maintaining relatively low computational cost, making them well suited to real-time and embedded systems, including many human-robot interaction scenarios. However, skeleton-based representations abstract away fine-grained motion information present in the underlying visual signal. While joint trajectories capture coarse movement patterns, they do not explicitly encode local motion dynamics such as subtle limb movements, object interactions, or inter-frame motion cues.

This limitation becomes particularly significant in early action anticipation, where partial observations may not yet exhibit distinctive pose configurations. In such cases, informative motion cues may be present even when pose alone is ambiguous, suggesting that augmenting skeleton-based representations with additional motion information is important for improving early prediction performance.

\subsection{Motion-based and Multimodal Approaches}
Motion-based representations, particularly those based on optical flow, capture dynamic information that complements pose and appearance features. Two-stream architectures \cite{simonyan2014TwoStreamConvolutionalNetworksActionRecognitionVideos} and more recent video models \cite{tong2022VideoMAEMaskedAutoencodersareDataEfficientLearnersSelfSupervisedVideoPreTraining} demonstrate the importance of motion cues for action understanding \cite{kong2022HumanActionRecognitionPredictionSurvey}. However, these approaches typically model motion as a dense, global signal, leading to substantial computational cost and redundancy. While multimodal fusion improves recognition performance, it does not explicitly structure motion representations around human kinematics. As a result, motion information is often encoded in a manner that is both computationally expensive and weakly aligned with human body dynamics, limiting its effectiveness for efficient and early action understanding.

Joint-centred motion representations, such as JOLO-GCN \cite{cai2021JOLOGCNMiningJointCenteredLightWeightInformationSkeletonBasedActionRecognition} and H-MoRe \cite{huang2025HMoReLearningHumancentricMotionRepresentationActionAnalysis}, have demonstrated that visual motion surrounding human body keypoints provides useful learnable information. Critically, however, these methods treat joint-centred motion as a separate, complementary input stream to skeleton keypoints rather than integrating motion features within the skeleton-based learning process itself. Moreover, both JOLO-GCN and H-MoRe are designed and evaluated specifically for action recognition, and have not been extended to action anticipation, where predictions must be made from partial, early observations of an unfolding action.

Across these approaches, a consistent trade-off emerges. Skeleton-based methods are efficient but lack fine-grained motion cues, while motion-based and multimodal methods capture richer dynamics at significantly higher computational cost. In interactive robotic systems, where perception must be both timely and resource-efficient, this trade-off becomes critical \cite{dragan2013Legibilitypredictabilityrobotmotion,villani2018SurveyhumanrobotcollaborationindustrialsettingsSafetyintuitiveinterfacesapplications}.

This reveals a clear gap in the literature. No existing work has sought to integrate joint-centred motion features directly within the skeleton-based action recognition framework and apply this to the task of action anticipation. Such integration would allow motion cues to interact with structural keypoint representations during learning, rather than being processed as a separate parallel signal, potentially yielding representations that are simultaneously more informative, better aligned with human body dynamics, and capable of supporting earlier and more accurate action predictions.

\section{Method}
\label{sec:method}

We propose PoseOFF, a pose-conditioned motion representation designed to encode localised motion dynamics aligned with human body structure. Rather than modelling motion as a dense global signal or fusing it with pose at the feature level, PoseOFF explicitly conditions motion representation on human pose, capturing motion cues at semantically meaningful body locations.

This design is motivated by the need for efficient and early action understanding in interactive robotic systems, where informative motion cues are often localised and temporally sparse. By structuring motion features around human kinematics, PoseOFF provides a compact and behaviourally grounded representation that supports action anticipation while avoiding the computational overhead of dense motion modelling.

By embedding early motion cues as learnable features in skeleton-based action recognition architectures, we aim to show such features can support state-of-the-art models achieving comparable accuracy with fewer observed frames.
Moreover, we seek to achieve higher action anticipation ability with minimal additional computational overheads, thus retaining the feasibility of the methods applied in real-time settings on standard computing hardware.

\subsection{PoseOFF Representation}
\label{subsec:PoseOFF}

PoseOFF encodes motion information by conditioning optical flow features on human pose. For each detected skeleton, motion features are extracted at the spatial locations of body joints, producing a representation that explicitly aligns motion dynamics with human kinematics.

Unlike conventional optical flow representations, which model motion densely across the entire frame, PoseOFF focuses on local regions associated with human movement. This results in a structured motion representation that captures behaviourally relevant dynamics while reducing redundancy.

The input to the PoseOFF representation is an RGB video consisting of $T$ frames. For each frame $I_t$, where $t \in [1, T]$, up to $M$ human skeletons are extracted, each defined by $V$ keypoints. Each keypoint is represented by $C_{\text{pose}} = 3$ channels, corresponding to either $(x,y,z)$ coordinates or $(x,y,\alpha)$, where $\alpha$ denotes pose estimation confidence.

To incorporate motion information, optical flow is computed between consecutive frames $I_t$ and $I_{t+1}$, yielding $T-1$ flow fields with $(u,v)$ vectors describing pixel-wise motion. Rather than representing motion densely across the entire frame, PoseOFF constructs motion features conditioned on the spatial configuration of the skeleton.

\begin{figure}[t]
  \centering
  \includegraphics[width=0.8\linewidth]{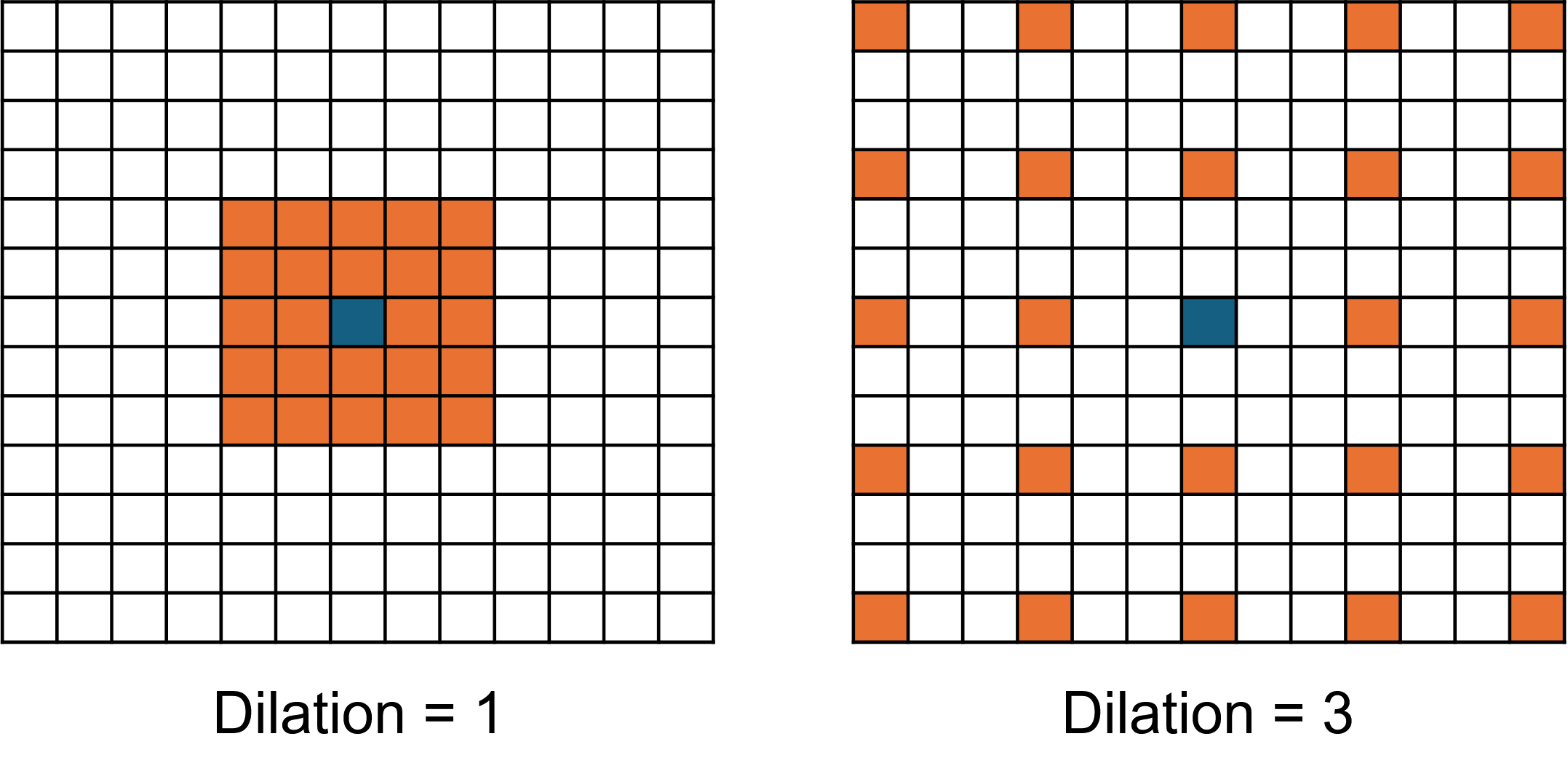}
  \caption{\label{fig:dilations}Example of $5\times5$ optical flow sampling windows with dilation 1 (left) and 3 (right). The centre cell corresponds to the pose keypoint location..}
\end{figure}

Specifically, for each keypoint location $(x_t, y_t)$, a local neighbourhood of optical flow vectors is extracted, forming an $N \times N$ region of motion centred at the joint. This produces a local motion feature of shape $(N, N, 2)$, capturing the motion dynamics in the vicinity of each body part. A dilation factor may be applied to control the spatial extent of this region, allowing the representation to capture motion at different scales (see~\autoref{fig:dilations}).

Each local motion region is flattened to produce $C_{\text{flow}} = 2N^2$ channels, which are concatenated with the corresponding pose features. This results in a unified, pose-conditioned representation with $C = C_{\text{pose}} + C_{\text{flow}} = 2N^2 + 3$ channels per keypoint.

The final PoseOFF representation for a sequence is:
\[
(T, M, V, C)
\]
where motion features are explicitly aligned with  joint locations, yielding a structured encoding of local motion dynamics.

\begin{figure}[t]
  \centering
  \includegraphics[width=0.6\linewidth]{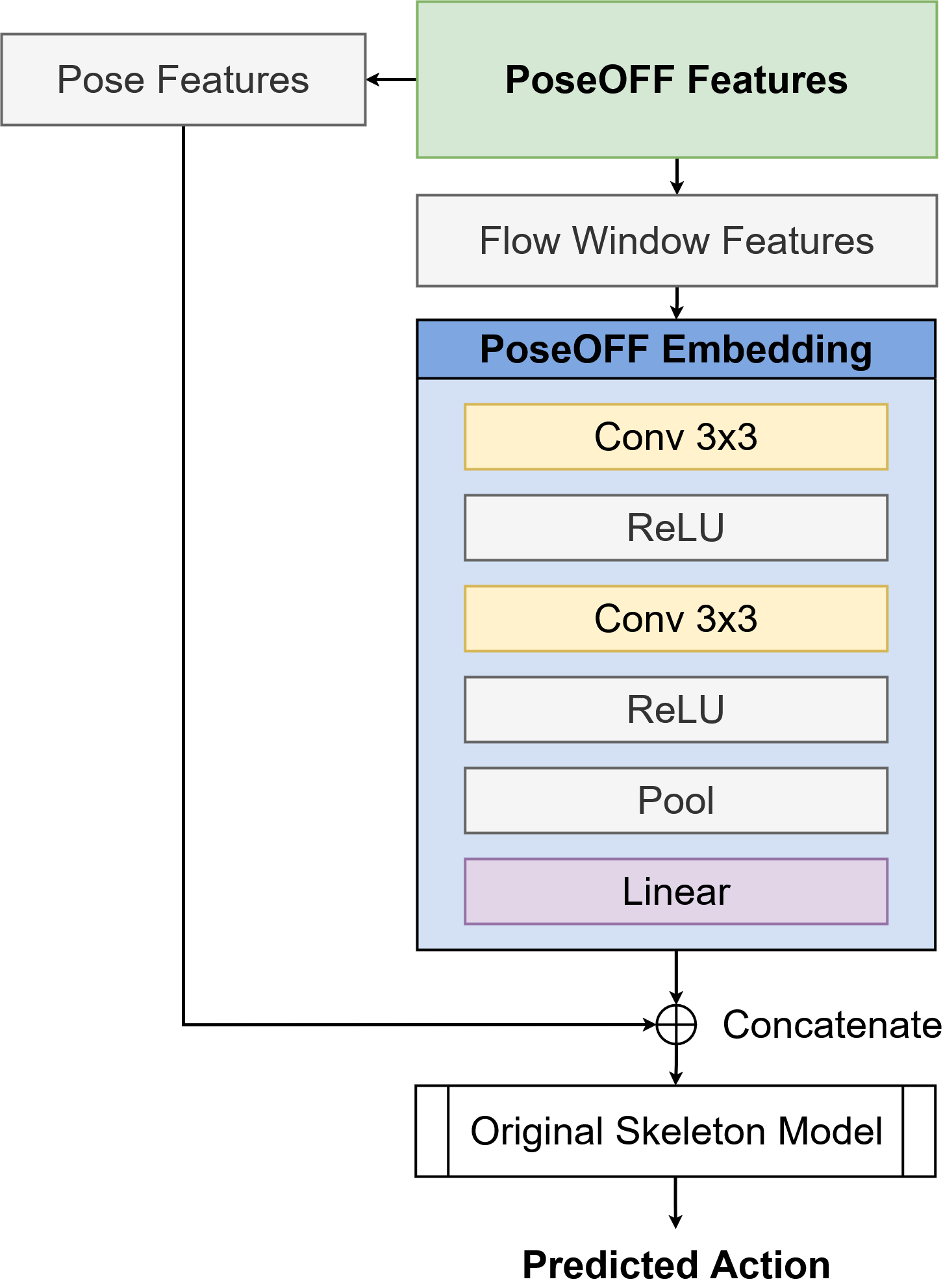}
  \caption{\label{fig:embedding_network_diagram}PoseOFF embedding layer diagram - convolutional layers (Conv) learn from optical flow vector windows, then concatenate these embeddings with the original pose features (x,y,z positions).}
\end{figure}

\subsection{Skeleton-based Action Recognition Models}
\label{subsec:models}
To demonstrate the applicability and effectiveness of PoseOFF, we modify three state-of-the-art skeleton keypoint action recognition models to incorporate learnable early visual motion cues.
The embedding layer of each model was modified to include a simple CNN designed to learn the features present in the windows of optical flow.
The CNN consisted of two convolutional layers separated by activation layers, pooling and a linear layer.
The output of these layers was concatenated with the original skeleton keypoints coordinates as additional channels in the skeleton graph.
A diagram of the PoseOFF embedding layer added to each model is presented in \autoref{fig:embedding_network_diagram}.

The models evaluated were InfoGCN++~\cite{chi2025InfoGCNLearningRepresentationPredictingFutureOnlineSkeletonBasedActionRecognition}, MS-G3D~\cite{liu2020DisentanglingUnifyingGraphConvolutionsSkeletonBasedActionRecognition}, and ST-GCN++~\cite{duan2022PYSKLGoodPracticesSkeletonActionRecognition}, trained using hyperparametes specified by the original authors and settings as close to those reported in the respective papers.

The PoseOFF extraction method utilises human skeleton \textit{joints} as locations in image space from which to extract windows of optical flow.
For this reason, the results presented in this paper are models trained and evaluated using the \textit{skeleton joint stream only}.

\subsubsectionmark{Observation Ratio}\label{subsubsec:obs_ratio}
To evaluate early action recognition, we employ temporal masking to simulate partial observation of input sequences. Specifically, only the initial portion of each sequence is provided to the model, requiring classification based on incomplete action execution. Sequences are masked according to predefined \textit{observation ratios}, which specify the proportion of frames retained as input. \autoref{fig:poseoff_extraction} illustrates an example with a 60\% observation ratio, where only the first part of the sequence is visible to the model. Partially observed sequences are padded to the required input length using the strategies defined in the original works: last frame duplication~\cite{chi2025InfoGCNLearningRepresentationPredictingFutureOnlineSkeletonBasedActionRecognition}, sequence replay~\cite{liu2020DisentanglingUnifyingGraphConvolutionsSkeletonBasedActionRecognition}, and zero padding~\cite{duan2022PYSKLGoodPracticesSkeletonActionRecognition}.

\subsection{Implementation}
\label{subsec:implementation}

All models were implemented in Python using the PyTorch framework \cite{paszkePyTorchImperativeStyleHighPerformanceDeepLearningLibrary}, building on publicly available code released by the respective authors.

Optical flow was computed using the RAFT algorithm \cite{teed2020RAFTRecurrentAllPairsFieldTransformsOpticalFlow}, with pretrained weights provided in the PyTorch ecosystem. Human pose keypoints were obtained directly from dataset annotations for NTU RGB+D 60 and NTU RGB+D 120. For the UCF101 dataset, pose keypoints were estimated using the YOLO-POSE Large model \cite{maji2022YOLOPoseEnhancingYOLOMultiPersonPoseEstimationUsingObjectKeypointSimilarityLoss}.

Training configurations were aligned with those reported in the original model implementations. For NTU RGB+D datasets, hyperparameters were kept consistent with the respective baseline settings. For UCF101, where no standard configuration is defined, hyperparameters were selected to closely match those used for NTU RGB+D 60. In all cases, baseline and PoseOFF-augmented models were trained using identical settings to ensure fair comparison.

\subsection{Datasets}
\label{subsec:datasets}
The models were trained and evaluated on three different datasets; the NTU RGB+D 60 dataset~\cite{shahroudy2016NTURGB+DLargeScaleDataset3DHumanActivityAnalysis}, NTU RGB+D 120 dataset~\cite{liu2020NTURGB+D120LargeScaleBenchmark3DHumanActivityUnderstanding} and the UCF101 dataset~\cite{soomro2012UCF101dataset101humanactionsclassesvideoswild}.

The NTU RGB+D 60 and NTU RGB+D 120 datasets were captured using three Microsoft Kinect v2 sensors and provide ground-truth 3D human skeleton data.
Each sample contains RGB videos, depth and infrared sequences, and 3D coordinates of 25 body joints.
Standard train/test splits defined by the authors ensure fair and consistent evaluation.
Samples are annotated with 60 and 120 action classes, respectively.
Owing to their widespread use in skeleton-based action recognition research ~\cite{chi2025InfoGCNLearningRepresentationPredictingFutureOnlineSkeletonBasedActionRecognition, duan2022PYSKLGoodPracticesSkeletonActionRecognition, liu2020DisentanglingUnifyingGraphConvolutionsSkeletonBasedActionRecognition}, these datasets were selected to establish baseline performance and to evaluate the impact of incorporating PoseOFF embeddings.

The UCF101 dataset consists of unconstrained, real-world videos depicting human actions.
Skeleton keypoints and optical flow data were estimated using the methods described in Sec.~\ref{subsec:implementation}.
The dataset includes three standard train/test splits.
UCF101 was chosen to assess performance under more challenging, ``in-the-wild'' conditions, in contrast to the controlled capture settings of the NTU RGB+D datasets.

The RGB video data of the NTU RGB+D datasets has a resolution of $1920\times1080$ pixels, whereas the UCF101 dataset has a resolution of $320\times240$ pixels.
Windows of optical flow for the PoseOFF embedding were sampled with a dilation factor of 3 for the NTU RGB+D datasets and a dilation factor of 1 for the UCF101 datasets.
A higher dilation factor was chosen for the NTU RGB+D datasets given their higher video resolution.
See~\autoref{fig:dilations} an example of dilated window sampling.

\section{Experiments}%
\label{sec:experiments}

\subsection{Evaluation Metrics}
For action classification, performance is measured using classification accuracy, defined as the percentage of samples for which the correct class label is predicted after observing the full action sequence.
For action anticipation, we report classification accuracy at each observation ratio as well as the Area Under the Curve (AUC). The AUC summarizes performance across all evaluated observation ratios and is computed as:

\[ \text{AUC}=\frac{\sum{\text{Acc}}_{\text{obs}}}{\text{N}_{\text{obs}}}\]
where $\text{Acc}_{\text{obs}}$ denotes the accuracy at a given observation ratio, and $\text{N}_{\text{obs}}$ is the number of evaluated observation ratios.

\subsection{Action Classification}%
\label{subsec:exp_action_classification}
To evaluate action classification performance, both the base models and their PoseOFF-augmented variants (\autoref{fig:embedding_network_diagram}) were trained and tested on full input sequences without temporal masking, simulating an offline action recognition setting.

\begin{table*}[b]
\centering
\caption{Classification accuracy (\%) for baseline and PoseOFF-augmented models across NTU RGB+D 60 (NTU), NTU RGB+D 120 (NTU120), and UCF101 datasets (joint stream only). Best results per model are in bold.}
\label{table:classification_accuracy}

\setlength{\tabcolsep}{5pt}

\begin{tabular}{llccccccc}
\toprule
 &  & \multicolumn{2}{c}{\textbf{NTU}}
    & \multicolumn{2}{c}{\textbf{NTU120}}
    & \multicolumn{3}{c}{\textbf{UCF101}} \\
\cmidrule(lr){3-4} \cmidrule(lr){5-6} \cmidrule(lr){7-9}
\textbf{Model} & \textbf{Variant}
    & \textbf{CS} & \textbf{CV}
    & \textbf{CSub} & \textbf{CSet}
    & \textbf{Split 1} & \textbf{Split 2} & \textbf{Split 3} \\
\midrule

\multirow{2}{*}{InfoGCN++}
    & Base
    & 83.52 & 91.38 & 75.92 & 77.61 & 59.21 & 57.82 & 58.90 \\
    & PoseOFF
    & \textbf{88.89} & \textbf{93.68} & \textbf{82.70} & \textbf{83.70}
    & \textbf{62.81} & \textbf{62.54} & \textbf{62.84} \\

\addlinespace

\multirow{2}{*}{MS-G3D}
    & Base
    & 87.40 & 93.43 & 76.24 & 78.75 & 61.39 & 62.02 & 63.45 \\
    & PoseOFF
    & \textbf{89.95} & \textbf{94.12} & \textbf{83.22} & \textbf{85.49}
    & \textbf{68.84} & \textbf{69.88} & \textbf{66.36} \\

\addlinespace

\multirow{2}{*}{ST-GCN++}
    & Base
    & 84.78 & 90.85 & 76.54 & 76.40 & 47.37 & 47.96 & 46.86 \\
    & PoseOFF
    & \textbf{88.88} & \textbf{93.70} & \textbf{82.36} & \textbf{85.00}
    & \textbf{58.27} & \textbf{60.13} & \textbf{58.61} \\

\bottomrule
\end{tabular}
\end{table*}

\begin{figure*}
  \centering
  \includegraphics[width=0.95\linewidth]{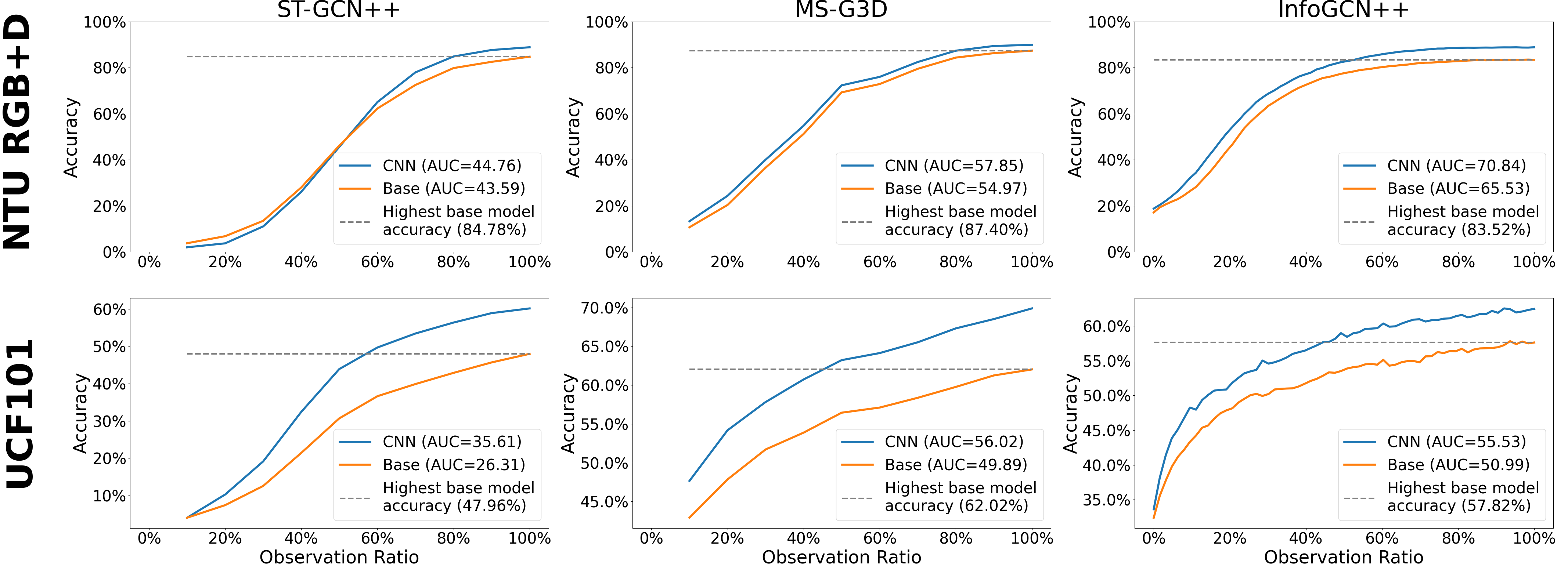}
  \caption{\label{fig:res_obs_ratio}Models trained using 100\% observation ratios, evaluated at 10\% observation ratio intervals on the NTU RGB+D 60 dataset, Cross Subject evaluation. The Area Under the Curve (AUC) is reported for each model in the sub figure labels.}
\end{figure*}

Classification results are reported in~\autoref{table:classification_accuracy}, with the best-performing model for each dataset and evaluation highlighted.
Across all architectures, incorporating PoseOFF embeddings yields average accuracy improvements of 2.14\%, 6.84\%, and 11.22\% on NTU RGB+D 60, NTU RGB+D 120, and UCF101, respectively,
The improvements on UCF101 are particularly notable, as this dataset consists of RGB-only videos captured “in the wild,” without depth or skeleton annotations.
The skeletons extracted with YOLO-POSE can occasionally be misplaced or dropped due to the unconstrained nature of the videos in the dataset, and given the aggressively abstract nature of skeleton keypoints, misplaced keypoints can significantly impact action recognition performance.
The addition of PoseOFF embeddings provides salient motion information and additional scene context, even when skeleton keypoints are misplaced.

\subsection{Action Anticipation}%
\label{subsec:exp_action_anticipation}

To evaluate whether PoseOFF enables earlier action prediction, temporal masking was used to simulate partial observation of ongoing actions. MS-G3D and ST-GCN++ were trained on full sequences (100\% observation) and evaluated at 10\% observation ratio increments. InfoGCN++ natively performs frame-level prediction over a fixed 64-frame input, and is therefore evaluated directly without interval-based masking~\cite{chi2025InfoGCNLearningRepresentationPredictingFutureOnlineSkeletonBasedActionRecognition}.

Results are shown in Fig.~\ref{fig:res_obs_ratio} for NTU RGB+D 60 (Cross-Subject) and UCF101. Performance is reported as classification accuracy at increasing observation ratios, with the baseline full-sequence accuracy indicated for reference. The point at which PoseOFF matches this baseline indicates how early equivalent recognition performance is achieved.

Across datasets, PoseOFF consistently enables models to reach baseline accuracy with fewer observed frames. On NTU RGB+D 60, ST-GCN++ and MS-G3D require only 80\% of the sequence to match full-sequence baseline performance, while InfoGCN++ achieves equivalent accuracy after observing 50\% of the input. Similar trends are observed on UCF101, where PoseOFF-augmented models match baseline performance after approximately half of the sequence.

These results demonstrate that PoseOFF improves early prediction capability, enabling comparable recognition performance under partial observation.

\subsection{Per-Class Performance Analysis}
\label{subsec:per_class}

To analyse where PoseOFF provides the greatest benefit, we compute per-class recall differences between baseline and PoseOFF models. Predictions are aggregated across models and confusion matrices are used to derive recall per class, providing a model-agnostic view of performance changes.

\begin{figure}[t]
    \centering

    \begin{subfigure}[t]{\linewidth}
        \centering
        \includegraphics[width=0.95\linewidth]{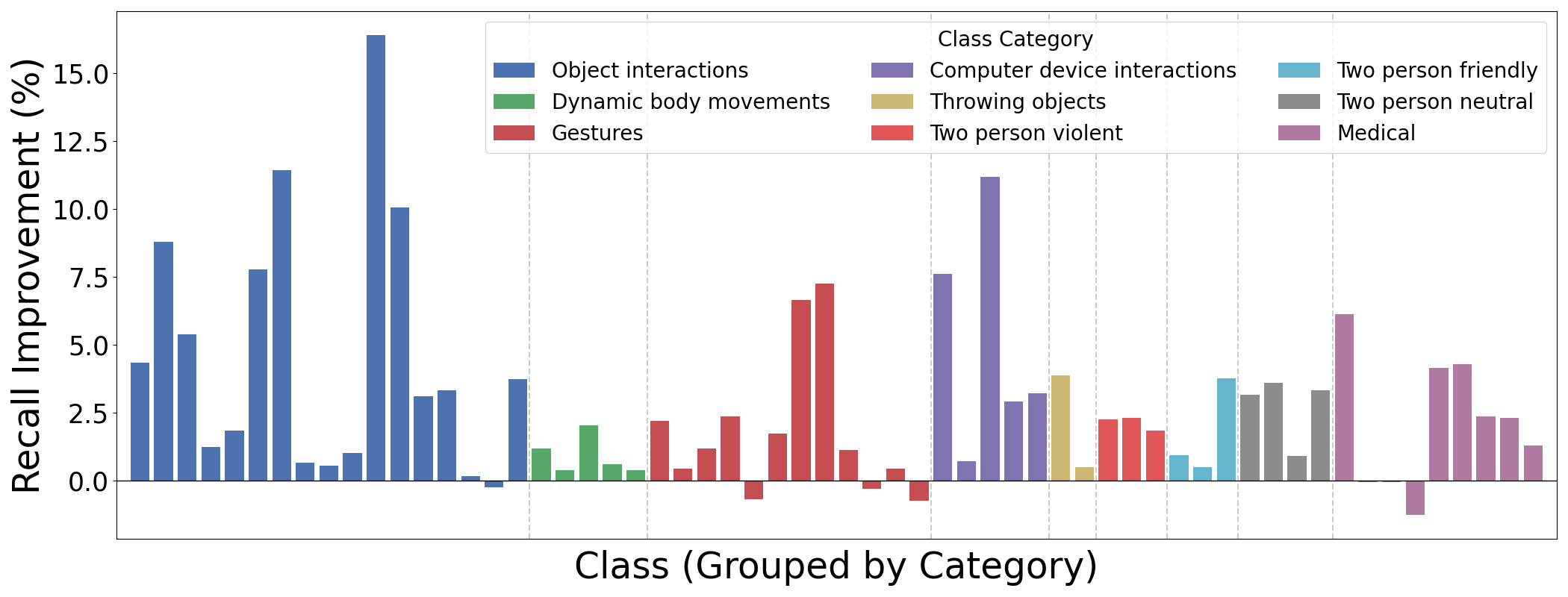}
        \caption{NTU RGB+D 60}
        \label{fig:perclass_ntu60}
    \end{subfigure}

    \vspace{0.5em}

    \begin{subfigure}[t]{\linewidth}
        \centering
        \includegraphics[width=0.95\linewidth]{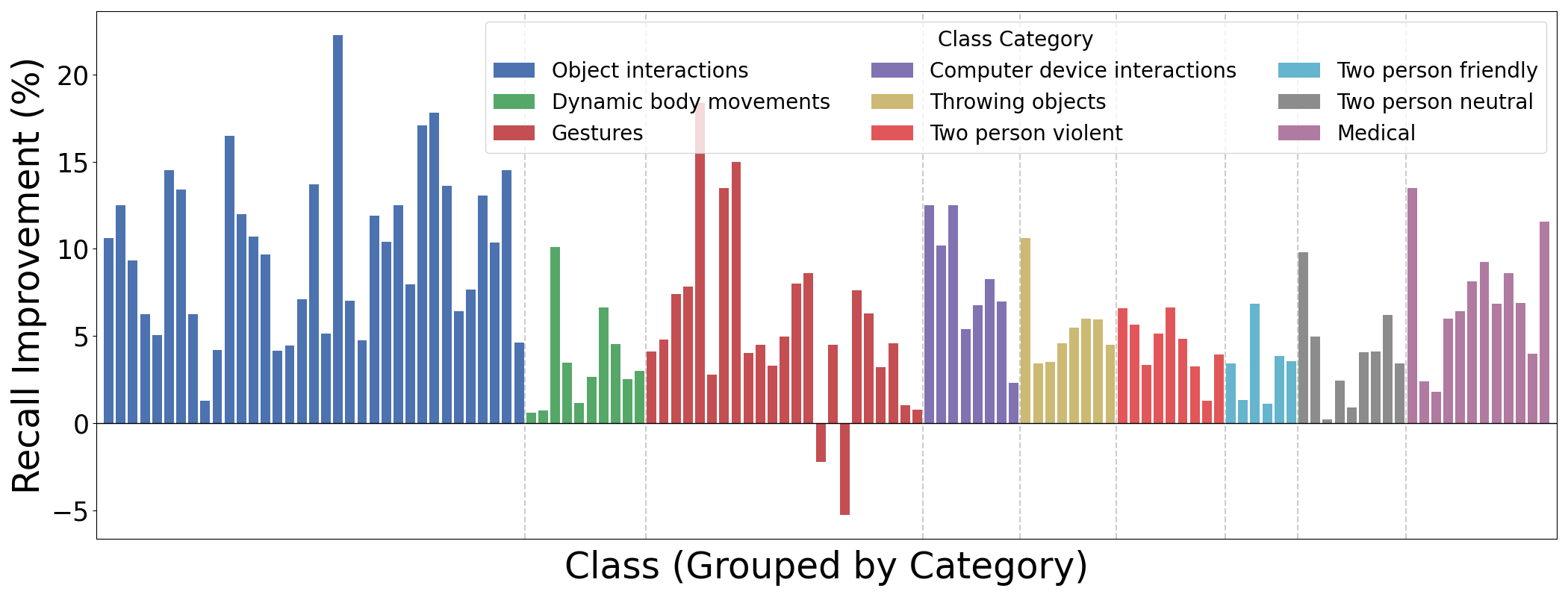}
        \caption{NTU RGB+D 120}
        \label{fig:perclass_ntu120}
    \end{subfigure}

    \vspace{0.5em}

    \begin{subfigure}[t]{\linewidth}
        \centering
        \includegraphics[width=0.95\linewidth]{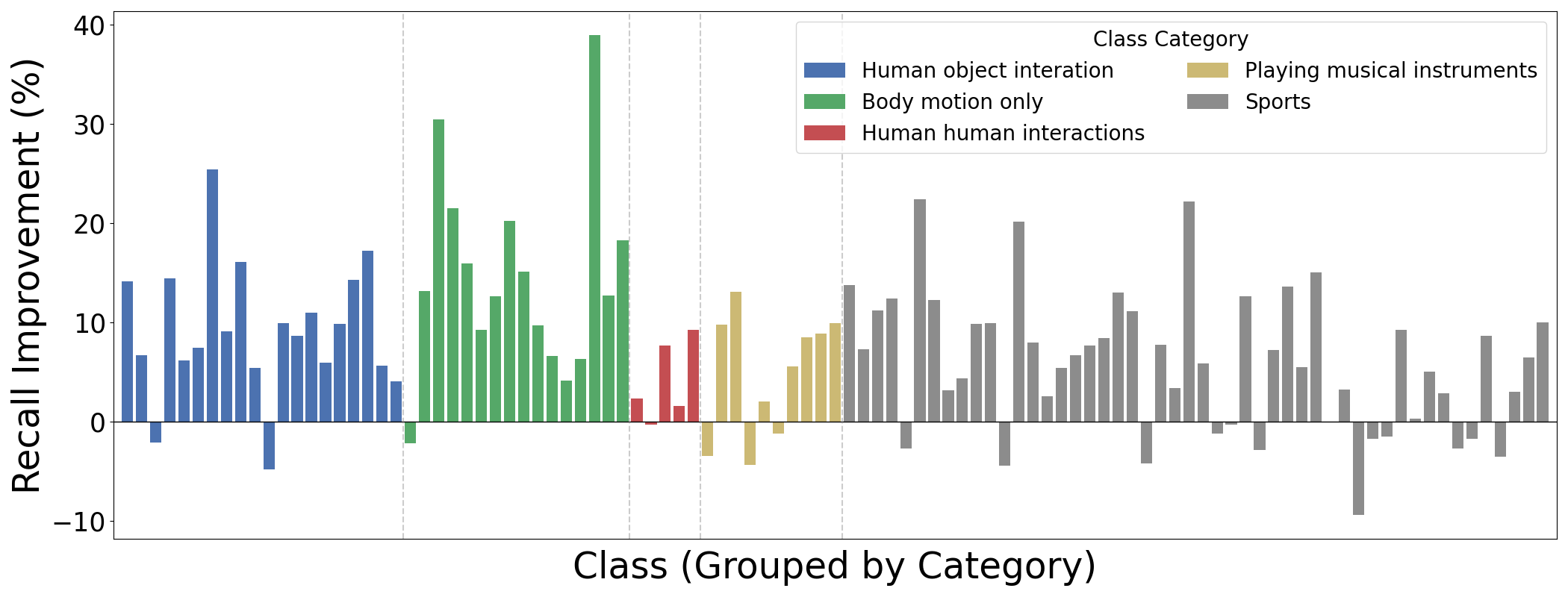}
        \caption{UCF101}
        \label{fig:perclass_ucf101}
    \end{subfigure}

    \caption{Per-class recall improvement (PoseOFF vs baseline) for (a) NTU RGB+D 60, (b) NTU RGB+D 120, and (c) UCF101. Classes are grouped by action category.}
    \label{fig:perclass_all}
\end{figure}

Per-class recall differences for each dataset are visualised in Fig.~\ref{fig:perclass_ntu60}, Fig.~\ref{fig:perclass_ntu120}, and Fig.~\ref{fig:perclass_ucf101}. Across all datasets, PoseOFF yields consistent improvements across the majority of action classes. On NTU RGB+D 60, performance improves for 88.33\% of classes, with a median recall gain of 2.12\% ($p < 10^{-9}$). On NTU RGB+D 120, improvements are observed for 98.33\% of classes, with a median gain of 5.98\% ($p < 10^{-20}$). On UCF101, PoseOFF improves 80.20\% of classes with a median gain of 7.65\% ($p < 10^{-13}$). These results indicate that gains are broadly distributed rather than concentrated in a small subset of classes.

The distribution of improvements reveals clear trends across action categories. On NTU datasets (Fig.~\ref{fig:perclass_ntu60}, Fig.~\ref{fig:perclass_ntu120}), the largest gains are observed in daily actions involving fine-grained and localised motion, particularly hand-object interactions and self-directed actions (e.g., writing, typing, applying cream, cutting nails). These actions are characterised by subtle, local motion cues that are not fully captured by skeletal pose alone.

A similar pattern is observed on UCF101 (Fig.~\ref{fig:perclass_ucf101}), where improvements are strongest in categories involving structured body motion and object interaction, while gains are more variable for sports-related actions. In particular, sports classes—characterised by large-scale, full-body motion—exhibit both strong improvements and occasional performance drops, suggesting that global motion dynamics may not always be fully captured by localised representations.

\begin{figure*}[t]
    \centering
    \includegraphics[width=0.95\linewidth]{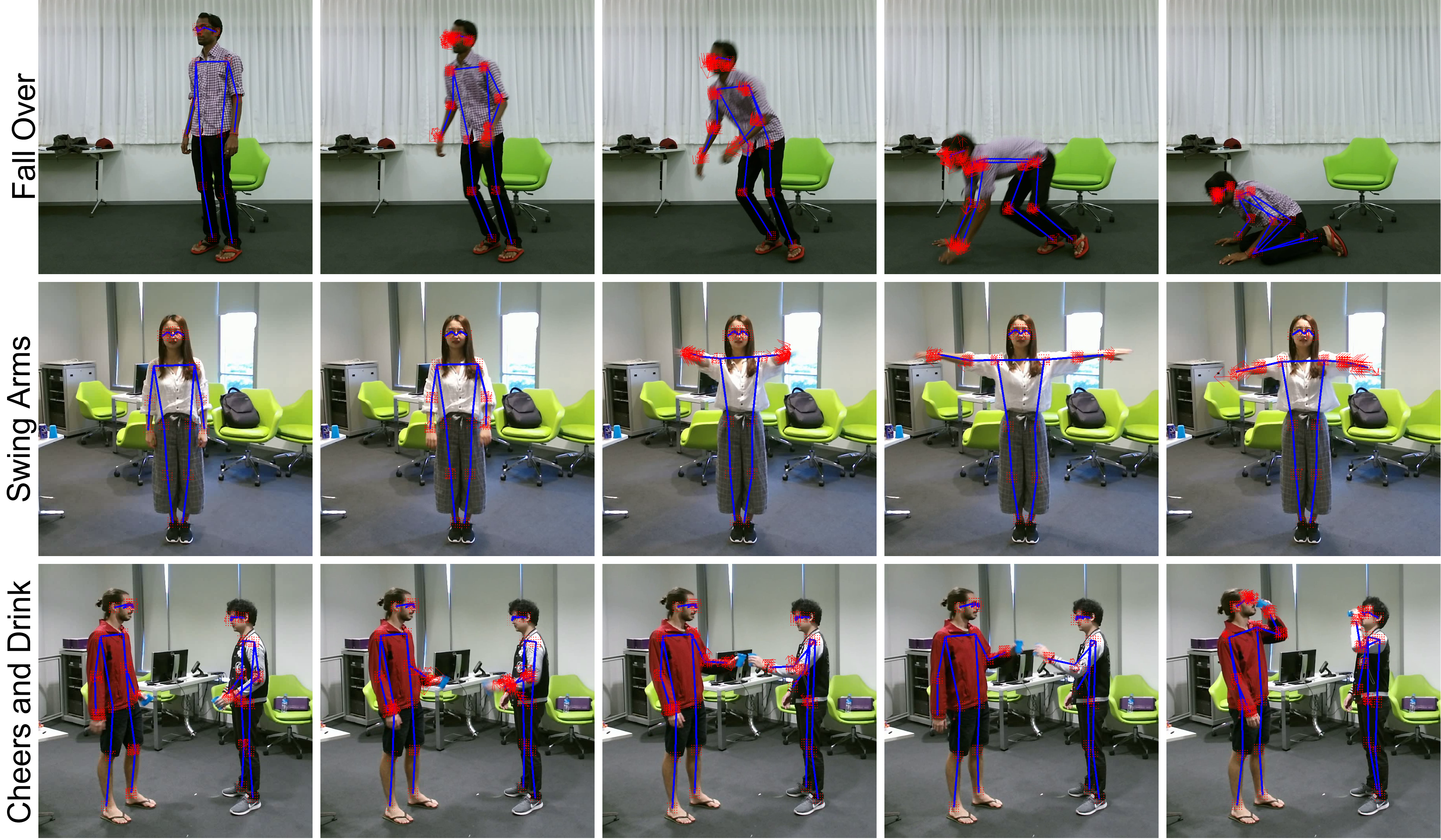}
    \caption{Qualitative examples illustrating how PoseOFF captures localised motion cues (e.g., rotation, divergence) around key joints, supporting improved recognition under occlusion, subtle motion, and scene clutter.}
    \label{fig:class_examples_full}
\end{figure*}

To further interpret these trends, Fig.~\ref{fig:class_examples_full} presents qualitative examples of action classes where PoseOFF yields improved recall. These examples illustrate how localised motion cues captured by pose-anchored flow contribute to improved discrimination under challenging visual conditions.

For instance, in \textit{fall over}, limb occlusion limits the availability of optical flow for some joints; however, motion patches anchored to neighbouring visible limbs still capture informative dynamics, enabling correct classification. In \textit{swing arms}, improvements arise from early detection of hand and elbow motion, which precedes significant changes in skeleton joint positions and would otherwise be weakly represented in pose-only features. In \textit{cheers and drink}, PoseOFF captures rotational motion of the elbows, although the example also highlights the presence of noisy flow patches in visually cluttered scenes.

Across these examples, the motion fields surrounding key joints (e.g., elbows and knees) exhibit structured patterns, including curl in the flow field corresponding to limb rotation, and divergence where joints move toward or away from the camera. These local motion signatures are not explicitly encoded in skeletal representations but are captured by PoseOFF.  Figure~\ref{fig:classexamples_small} provides two more examples (\emph{pick-up} and \emph{jump}) where PoseOFF yields improved recall.

Conversely, performance decreases are limited to a small number of classes across all datasets. These typically correspond to actions dominated by global motion patterns or complex multi-agent interactions, where localised motion cues alone may be insufficient for reliable discrimination. For the UCF101 dataset, actions where human body joints are partially occluded - like videos of action classes ``Rowing'' or ``Breast Stroke'' - pose-only models show greater action classification accuracy. This could be caused by an introduction of significant noise in the flow field, which has less of an impact on the pose keypoints.

\subsection{Performance and Implementation Considerations}
\label{subsec:inference_time}
PoseOFF provides a computationally efficient alternative to dense motion representations, retaining informative motion cues while avoiding the substantial overhead of full optical flow processing. Table~\ref{table:datasize_comparison} compares data size and extraction cost for pose-only, full optical flow, and PoseOFF representations. For a representative $1920\times1080$ video sequence, PoseOFF requires approximately 1060kB of data, compared to 60kB for pose-only and 1.66GB for full optical flow. This represents an orders-of-magnitude reduction in data relative to dense motion representations, while still incorporating meaningful motion information.

PoseOFF adds only a modest overhead to action inference time compared to pose-only inference, due to processing optical flow vectors embedded with keypoint features. As shown in Table~\ref{table:model_size_inference}, it increases trainable parameters by an average of 1.88\% and adds 5.3 ms and 9.35 ms to post-extraction inference time for InfoGCN++ and MS-G3D, respectively. These increases remain small relative to model complexity and are compatible with real-time operation.

Importantly, the modifications required to incorporate PoseOFF are limited to the embedding layer of each model, preserving the structure and efficiency of the underlying architecture. This enables straightforward integration into existing skeleton-based pipelines.

\begin{table}[b]
\centering
\caption{Data size and extraction speed comparison for pose-only, PoseOFF, and full optical flow representations.}
\label{table:datasize_comparison}
\begin{tabular}{lccc}
\toprule
\textbf{Representation} & \textbf{Data Size} & \textbf{Extraction Time} & \textbf{FPS} \\
\midrule
Pose only  & 60\,kB   & 713\,ms  & 140 \\
PoseOFF    & 1060\,kB & 720\,ms  & 138 \\
Full flow  & 1.66\,GB & 10.8\,s  & 9 \\
\bottomrule
\end{tabular}
\end{table}

\begin{table}[b]
\centering
\caption{Model size and inference speed comparison for baseline and PoseOFF-augmented models.}
\label{table:model_size_inference}
\begin{tabular}{l l cc}
\toprule
\textbf{Model} & \textbf{Variant} & \textbf{Parameters} & \textbf{Inference Time} \\
\midrule
\multirow{2}{*}{InfoGCN++}
    & Base    & 613,747  & 12.15\,ms (82 fps) \\
    & PoseOFF & 620,448  & 17.44\,ms (57 fps) \\

\addlinespace

\multirow{2}{*}{MS-G3D}
    & Base    & 2,194,476 & 13.13\,ms (76 fps) \\
    & PoseOFF & 2,204,420 & 22.48\,ms (44 fps) \\

\addlinespace

\multirow{2}{*}{ST-GCN++}
    & Base    & 516,454  & 4.43\,ms (226 fps) \\
    & PoseOFF & 537,648  & 4.20\,ms (238 fps) \\

\bottomrule
\end{tabular}
\end{table}

\section{Discussion and Conclusion}
\label{sec:discussion_conclusion}

The results demonstrate that incorporating localised motion cues through PoseOFF consistently improves early action recognition, particularly under partial observation. By conditioning motion extraction on pose, PoseOFF captures fine-grained dynamics—such as limb acceleration and object interaction—that are not represented in skeletal configurations alone, while avoiding the overhead of dense optical flow. This positions PoseOFF as an efficient middle ground between lightweight pose-based methods \cite{yan2018SpatialTemporalGraphConvolutionalNetworksSkeletonBasedActionRecognition, shi2019SkeletonBasedActionRecognitionDirectedGraphNeuralNetworks} and computationally intensive motion-based approaches \cite{simonyan2014TwoStreamConvolutionalNetworksActionRecognitionVideos}.

Per-class analysis shows that these gains are most pronounced for actions characterised by subtle, localised motion, including hand-object interactions and self-directed activities, where pose trajectories alone are insufficient for early discrimination. In contrast, actions dominated by large-scale global motion exhibit smaller or more variable improvements, suggesting that such dynamics are already well captured by skeletal representations.

From a systems perspective, PoseOFF provides a lightweight mechanism for enhancing existing skeleton-based models with targeted motion cues, requiring minimal architectural modification and modest computational overhead. This makes it well suited to low-latency and resource-constrained settings.

These properties are particularly relevant in human-robot interaction scenarios, where early and reliable interpretation of human intent directly impacts responsiveness, coordination, and safety \cite{dragan2013Legibilitypredictabilityrobotmotion, lasota2017SurveyMethodsSafeHumanRobotInteraction}. By enabling comparable performance with fewer observed frames, PoseOFF supports earlier prediction of human actions, facilitating more anticipatory behaviours such as motion priming, adaptive planning, and proactive safety responses.

PoseOFF relies on the quality of pose estimation and local motion extraction, and may be less effective under severe occlusion or highly dynamic global motion. Future work will explore more efficient motion estimation strategies and real-time deployment to evaluate system-level performance in embodied interaction settings.

\bibliographystyle{IEEEtran}  
\bibliography{refs}

\end{document}